\documentclass[fleqn,12pt]{wlscirep}

\usepackage[utf8]{inputenc}
\usepackage[T1]{fontenc}
\usepackage{titling}
\usepackage{bm}
\usepackage{setspace}
\usepackage{caption}
\usepackage{makecell}
\usepackage{graphicx}    
\usepackage{subcaption}  
\usepackage[mathlines]{lineno}
\usepackage{multirow}
\usepackage{booktabs}
\usepackage{url}
\usepackage{parskip}
\usepackage{soul}
\usepackage{orcidlink}

\usepackage{xcolor}
\usepackage{soul} 

\begin{document}

\begin{titlepage}
\raggedright
\textbf{Original Article} \\
\vspace{1.0cm}

\begin{center}
\LARGE\textbf{Retrieval-Grounded Multi-View Reasoning for Clinically Reliable Radiology Report Generation} \\
\vspace{1.2cm}
\normalsize
\small

\textbf{
Wanshu Fan\textsuperscript{1}~\orcidlink{0000-0001-6299-2795},
Zeyu Wang\textsuperscript{1},
Zheng Tong\textsuperscript{1},
Qi Ai\textsuperscript{2},
Wencai Weng\textsuperscript{2},
Haifan Gong\textsuperscript{3},
Congyu Liao\textsuperscript{4},
Guoliang You\textsuperscript{5},
Xiaofeng Liu\textsuperscript{6},
Dongsheng Zhou\textsuperscript{1,*},
Cong Wang\textsuperscript{7,*}~\orcidlink{0000-0002-6068-0103}} \\
\vspace{0.3cm}

\textsuperscript{1} National and Local Joint Engineering Laboratory of Computer Aided Design, School of Software Engineering, Dalian University, Dalian, China\\
\textsuperscript{2}Department of Surgery, Affiliated Zhongshan Hospital of Dalian University, Dalian, China\\
\textsuperscript{3}The Chinese University of Hong Kong, Shenzhen, China\\
\textsuperscript{4}University of California, San Francisco, USA\\
\textsuperscript{5}School of Computer Science and Technology, University of Science and Technology of China, Hefei, China\\
\textsuperscript{6}Yale University, New Haven, USA\\
\textsuperscript{7}The Hong Kong Polytechnic University, Hong Kong, China
\end{center}

\vspace{0.3cm}
{Correspondence and requests for materials should be addressed to:\\
    Dongsheng Zhou: zhouds@dlu.edu.cn   \\
Cong Wang: supercong94@gmail.com}

\vspace{0.2cm}

\vfill
\end{titlepage}

\newpage
\raggedright
\textbf{\large Abstract}
\vspace{1cm}

\raggedright

Chest radiograph report generation is hampered by retrieved reports that resemble the current image but are clinically inconsistent with it, and by diagnostic-label imbalance. We present RG-MVR, a retrieval-grounded framework comprising Retrieval-Augmented Evidence Grounding, which combines dense memory summaries with visually gated sparse retrieved evidence; Adaptive Multi-view Evidential Reasoning, which weights image, memory, and retrieved evidence per case; and a Clinical Prompt Learner, which models finding mention separately from positive, negative, or uncertain status. On MIMIC-CXR, RG-MVR achieved a clinical-label F1 of 0.483 and a BLEU-4 of 0.118 (IU X-ray: 0.202 and 0.110). In an exploratory external cohort of 22 examinations, clinical-label agreement declined markedly (F1, 0.071) and unsupported statements remained common, although physicians found fewer unsupported temporal comparisons (9 vs.\ 20 cases) and device-related statements (7 vs.\ 15) than with the baseline. Evidence-aware retrieval and structured diagnostic prompting may reduce specific reporting errors, but broader external validation is required.

\vspace{2cm}

\newpage

\setstretch{1.2}

\section{Introduction}

Chest radiography is among the most frequently performed imaging examinations in clinical practice, and the free-text radiology report remains the principal means by which a radiologist's interpretation becomes an actionable clinical record. Radiology report generation (RRG) seeks to automate part of this process by producing clinically meaningful free-text drafts from images, which remain subject to radiologist oversight~\cite{PengFSLYZWZ24}.
Unlike conventional image captioning, RRG requires faithful grounding of subtle abnormalities, precise expression of diagnostic uncertainty, and a coherent narrative structure, because errors at any of these levels can compromise downstream clinical decisions.
These requirements are particularly demanding for chest radiographs, in which normal regions dominate, findings are severely imbalanced~\cite{2026WZY}, and the same finding can be described in many ways. As a result, existing systems often produce reports that are fluent yet clinically incomplete: they over-generate frequent, template-like statements (e.g., ``no acute cardiopulmonary process'') and overlook rare but clinically significant abnormalities~\cite{2024cfh}.

Recent work has advanced RRG along two main directions. First, knowledge-injected architectures have improved cross-modal visual--textual interaction~\cite{huang2023kiut}, and attention-guided decoding and state-space modeling~\cite{hou2025rrg} have strengthened long-range coherence in generated narratives~\cite{lang2025dacg}. Second, retrieval-enhanced frameworks and diagnosis-driven prompting~\cite{jin2024promptmrg} use historical reports or predicted disease labels as auxiliary context, steering generation toward clinically plausible content and mitigating errors on rare findings~\cite{zhou2025learnable,xu2025deep}. These approaches have improved both linguistic fidelity and clinical efficacy, as measured by structured-label extraction tools, and they motivate the retrieval-augmented design adopted here.

Despite these advances, three interrelated limitations remain.
First, embedding similarity alone does not guarantee the clinical relevance of retrieved reports. A report retrieved from a shared image--text representation space may resemble the query image yet describe findings that are absent from, or inconsistent with, the current examination; without effective evidence selection and refinement, such mismatches can introduce irrelevant or unsupported statements into the generated report~\cite{zhou2025learnable}.
Second, existing fusion strategies may not account for case-specific differences in the relevance of visual features and retrieved text. A subtle, rare finding may benefit from stronger reliance on retrieval-derived priors, whereas a case with an unambiguous visual signal should rely predominantly on the image; rigid fusion can therefore either propagate irrelevant retrieved content or suppress clinically important visual cues~\cite{huang2023kiut,lang2025dacg}.
Third, diagnosis-conditioned generation typically relies on a single multi-class prediction per finding, which does not distinguish whether a finding is mentioned from its diagnostic polarity (positive, negative, or uncertain). Given the strong imbalance of diagnostic states in chest radiograph datasets, this formulation tends to favor the dominant ``not mentioned'' state and to under-recognize infrequent positive findings~\cite{xu2025deep}. Modeling mention status and polarity separately may therefore provide more structured diagnostic guidance for report generation.

\begin{figure*}[t]
\centering
\includegraphics[width=1\textwidth]{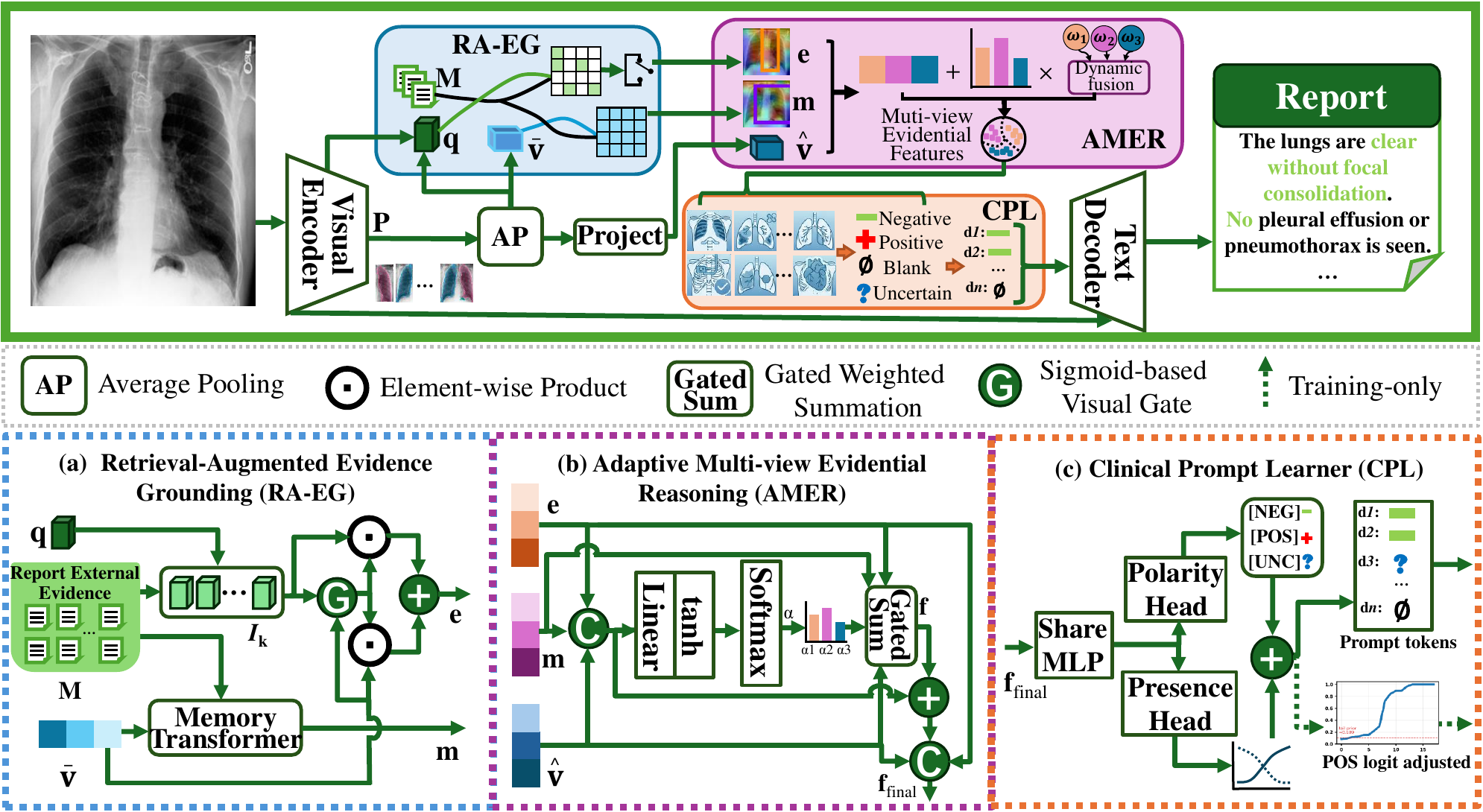}
\caption{Overview of the proposed RG-MVR framework.
The model combines Retrieval-Augmented Evidence Grounding (RA-EG), which summarizes dense memory and denoises sparse retrieved evidence;
Adaptive Multi-view Evidential Reasoning (AMER), which dynamically fuses visual, dense-memory, and sparse-evidence representations;
and the Clinical Prompt Learner (CPL), which converts predicted clinical states into decoder prompt tokens.}
\label{fig1:all}
\end{figure*}

To address these limitations, we propose Retrieval-Grounded Multi-View Reasoning (RG-MVR), a chest radiograph report generation framework that combines retrieval evidence refinement, adaptive multi-view fusion, and structured diagnostic prompting (Fig.~\ref{fig1:all}). Retrieval-Augmented Evidence Grounding (RA-EG) separates retrieved information into a dense memory summary and sparse top-ranked evidence, and uses visual gating to attenuate retrieved content that is poorly supported by the image. Adaptive Multi-view Evidential Reasoning (AMER) integrates visual features, the memory summary, and the refined evidence through case-adaptive weighting and residual refinement, allowing the contribution of each source to vary across examinations. The Clinical Prompt Learner (CPL) adopts a hierarchical formulation that separates whether a finding is mentioned from its diagnostic polarity and converts the predicted states into discrete prompt tokens that guide the decoder. The framework is trained end-to-end with language-generation and diagnostic-classification objectives balanced by learnable task weights.
Our contributions are summarized as follows:
\begin{itemize}
\item We propose RG-MVR, a retrieval-grounded, multi-view reasoning framework for chest radiograph report generation that explicitly models the reliability of retrieved evidence rather than treating it as uniformly trustworthy.
\item We design RA-EG, which combines dense memory summarization, sparse top-$k$ evidence aggregation, and visual-gated refinement to attenuate retrieved reports that are topically similar but clinically irrelevant.
\item We develop AMER, which re-weights visual, memory, and retrieved evidence for each case and refines the fused representation through a residual pathway for downstream diagnosis and generation.
\item We design CPL, which decouples finding mention from diagnostic polarity and converts the predicted states into discrete prompt tokens; the full model is trained with a learnable precision-weighted multi-task objective to cope with severe class imbalance.
\item 
We adopt a tiered evaluation that separates benchmark evidence from external and safety evidence: MIMIC-CXR serves as the primary benchmark, IU X-ray as an external cross-dataset validation set, and a local digital radiography (DR) cohort as an external generalization and failure-mode stress test with exploratory two-physician assessment.
\end{itemize}

\section{Results}

\begin{table}[t]
\centering
\caption{
Cohort characteristics and evaluation splits.   
MIMIC-CXR~\cite{johnson2019mimic} provides the primary training, validation, and test benchmark, whereas IU X-ray~\cite{DKR16} is used only for external cross-dataset validation. 
Demographic fields are available only for the linked MIMIC-CXR subset and are not reported for IU X-ray.}
\label{tab:cohort}
\scalebox{0.553}{
\begin{tabular}{ccc}
\toprule
 & MIMIC-CXR~\cite{johnson2019mimic} & IU X-ray~\cite{DKR16} \\
\midrule
Patients & 60{,}547 (this split) & Not reported \\
Studies & 227{,}835 & 3{,}955 (reports) \\
Images & 377{,}110 & 7{,}470 \\
Site & Beth Israel Deaconess Medical Center (single center) & Indiana Network for Patient Care (multi-institutional) \\
Collection period & 2011--2016 & Not reported \\
Sex distribution (this split) & 47.7\% male / 52.3\% female [n=57{,}270] & Not reported \\
Race/ethnicity (this split) & \makecell{White 63.4\%, Black/African American 17.1\%, Asian 3.6\%, Hispanic/Latino 6.4\%, \\ Other/Unknown 9.6\% [n=47{,}855]} & Not reported \\
Age distribution (this split) & Mean 56.9 yrs, median 58, range 18--91; 47.6\% $\geq$60 yrs [n=57{,}270] & Not reported \\
Train/Val/Test split (this study, samples\footnotemark) & 270{,}790 / 2{,}130 / 3{,}858 & \makecell{External validation only: 4{,}168 records, all used as test}
\\
\bottomrule
\end{tabular}}
\end{table}
\footnotetext{Sample counts follow the preprocessing of Chen et al. (2020), as reused by prior RRG work (e.g., PromptMRG); this split is not a study-level partition of the 227{,}835 studies reported above.}
\subsection{Study design}
We evaluated RG-MVR in a tiered design. MIMIC-CXR~\cite{johnson2019mimic} served as the primary benchmark for model comparison and ablation, using the 270{,}790/2{,}130/3{,}858 train/validation/test sample split adopted in prior RRG studies.
IU X-ray~\cite{DKR16} was used solely for external cross-dataset validation: MIMIC-CXR-trained weights were evaluated on all 4{,}168 records in the preprocessed IU X-ray file, without IU-specific training or validation.
In addition, a local cohort of 22 DR examinations was reserved for external generalization and safety stress testing. Because its size cannot support stable estimates of comparative performance, this cohort was not used as a primary benchmark; instead, it was used to characterize cross-domain degradation, clinician-identified error modes, and RadGraph-based semantic overlap.

\subsection{Patient characteristics}
Table~\ref{tab:cohort} summarizes the demographic and acquisition characteristics of the two public datasets. MIMIC-CXR comprises 377{,}110 chest radiographs from 227{,}835 imaging studies of 65{,}379 patients who presented to the Beth Israel Deaconess Medical Center Emergency Department between 2011 and 2016~\cite{johnson2019mimic}.
Our train/validation/test split of 270{,}790/2{,}130/3{,}858 samples follows the preprocessing of Chen et al.~\cite{chen-etal-2020-generating}, which has been reused by subsequent retrieval-augmented and diagnosis-driven RRG methods (e.g., PromptMRG~\cite{jin2024promptmrg}). This sample-level partition covers 60{,}547 unique patients and is distinct from, rather than a subdivision of, the 227{,}835 studies reported above.
%

Patient-level demographic attributes (self-reported sex, anchor age at first admission, and race/ethnicity recorded at the earliest admission) were obtained by linking MIMIC-CXR subjects to the MIMIC-IV clinical database.
Sex and age were available for 57{,}270 of the 60{,}547 patients in our split (94.6\%) and race/ethnicity for 47{,}855 (79.0\%); the remaining subject identifiers could not be matched to MIMIC-IV clinical records.
Among patients with available data, 52.3\% were female and 47.7\% male; the mean age was 56.9 years (median 58; range 18--91), and 47.6\% were aged 60 years or older. Most patients were White (63.4\%), followed by Black/African American (17.1\%), Hispanic/Latino (6.4\%), Asian (3.6\%), and other or unknown race/ethnicity (9.6\%). Compared with the training split, the test split was older (mean 67.7 vs.\ 56.8 years; 73.8\% vs.\ 47.5\% aged \(\ge\)60 years) and included a higher proportion of White patients (74.5\% vs.\ 63.3\%), which should be considered when interpreting test-set performance. The IU X-ray release comprises 3{,}955 de-identified reports and 7{,}470 associated frontal and lateral images collected through the Indiana Network for Patient Care~\cite{DKR16,seyyed2021underdiagnosis}; no age, sex, or race/ethnicity breakdown is publicly available for this cohort.

Two caveats apply to this cohort characterization. First, the MIMIC-CXR demographic statistics are computed from patients with available MIMIC-IV linkage (94.6\% for sex and age; 79.0\% for race/ethnicity) rather than from the full 60{,}547-patient split, and the 3{,}277 patients without sex and age data or the 12{,}692 patients without admission records may differ systematically from the analyzed subset. Second, the absence of public demographic metadata for IU X-ray precludes cohort characterization for that dataset.

\begin{figure*}[t]
\centering
\includegraphics[width=0.99\textwidth]{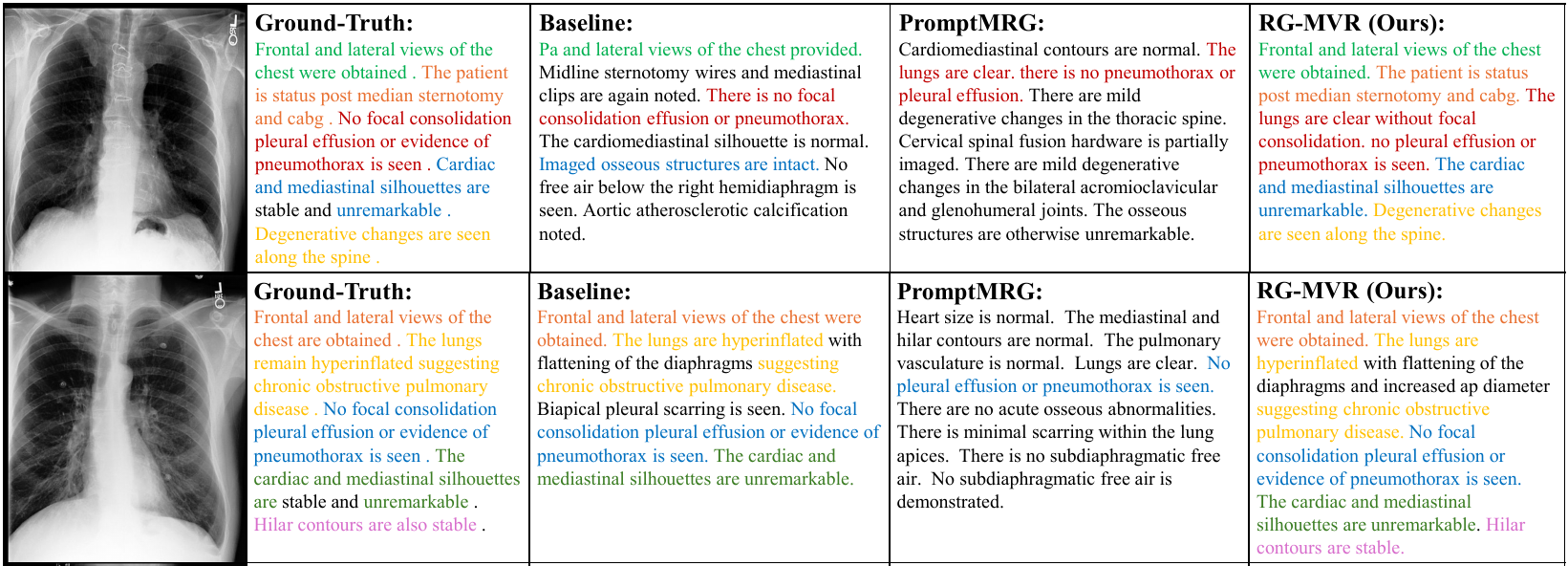}
\caption{Qualitative results on the MIMIC-CXR dataset. Matches with the ground truth are highlighted in the same colors, while inconsistencies are marked in black.}
\label{fig3:env}
\end{figure*}
\begin{table*}[t]
\centering
\caption{
Comparison with published report-generation methods on MIMIC-CXR~\cite{johnson2019mimic} and IU X-ray~\cite{DKR16}.
On MIMIC-CXR, RG-MVR achieves the best or tied-best results on most metrics, including all clinical-efficacy metrics.
On IU X-ray, RG-MVR leads on BLEU-1 to BLEU-4 and ROUGE-L, whereas its clinical-efficacy scores remain below those of REVTAF and PromptMRG.
NLG, natural language generation; CE, clinical efficacy (CheXbert-based precision, recall, and F1). \textbf{Bold} values indicate the best value within each dataset block; dashes indicate metrics not reported in the original publications.
}
    \footnotesize
    \setlength{\tabcolsep}{9pt} 
    \scalebox{0.75}{
\begin{tabular}{l|l|ccccccc|ccc}
\toprule
\multicolumn{1}{c|}{\multirow{2}{*}{\textbf{Data}}}                               & \multicolumn{1}{c|}{\multirow{2}{*}{\textbf{Model}}}  & \multicolumn{7}{c}{\textbf{NLG Metrics}}  & \multicolumn{3}{c}{CE Metrics}                    \\ \cmidrule{3-12}
\multicolumn{1}{c|}{\multirow{23}{*}{\textbf{\rotatebox{90}{MIMIC-CXR}}}}                   &                                                                                         & BLEU-1 & BLEU-2 & BLEU-3 & BLEU-4                            & ROUGE-L          &METEOR  &CIDEr    & Precision & Recall & F1            \\ \midrule 

                                                       & \multicolumn{11}{c}{\textit{\textbf{Early Image Captioning Methods}}}                                                                                                                                                                                       \\ \cmidrule{2-12}
                                                       & \multicolumn{1}{l|}{ST \cite{vinyals2015tell}}                       & 0.299                          & 0.184                          & 0.121                          & 0.084                          & 0.263                 & 0.124    &-    & 0.249 & 0.203 & 0.204  \\
                                                       & \multicolumn{1}{l|}{Att2In \cite{rennie2017selfcritical}}             & 0.325                          & 0.203                          & 0.136                          & 0.096                          & 0.276         & 0.134        &-      & 0.322 & 0.239 & 0.249    \\
                                                       & \multicolumn{1}{l|}{AdaAtt \cite{lu2017knowing}}                      & 0.299                          & 0.185                          & 0.124                          & 0.088                          & 0.266                  & 0.118     &-  & 0.268 & 0.186 & 0.181  \\
                                                       & \multicolumn{1}{l|}{TopDown \cite{anderson2018bottomup}}              & 0.317                          & 0.195                          & 0.130                          & 0.092                          & 0.267                 &-      &-  & 0.320 & 0.231 & 0.238  \\ \cmidrule{2-12} 
                                                       & \multicolumn{11}{c}{\textit{\textbf{From-Scratch RRG Methods}}}                                                                                                                                                                                             \\ \cmidrule{2-12}
                                                       & \multicolumn{1}{l|}{R2Gen \cite{chen-etal-2020-generating}}           & 0.353                          & 0.218                          & 0.145                          & 0.103                          & 0.277               & 0.142  &-        & 0.333 & 0.273 & 0.276 \\
                                                       & \multicolumn{1}{l|}{CA \cite{liu-etal-2021-contrastive}}                       & 0.350                          & 0.219                          & 0.152                          & 0.109                          & 0.283                 & 0.151     &-   & -     & -     & - \\
                                                       & \multicolumn{1}{l|}{CMCL \cite{liu-etal-2021-competence}}            & 0.344                          & 0.217                          & 0.140                          & 0.097                          & 0.281                       &0.133 &-  & -     & -     & - \\
                                                       & \multicolumn{1}{l|}{PPKED \cite{liu2021ppked}}                        & 0.360                          & 0.224                          & 0.149                          & 0.106                          & 0.284         & 0.149  &-            & -     & -     & -   \\
                                                       & \multicolumn{1}{l|}{R2GenCMN \cite{chen-etal-2021-cross-modal}}       & 0.353                          & 0.218                          & 0.148                          & 0.106                          & 0.278                   & 0.142   &-   & 0.334 & 0.275 & 0.278 \\
                                                       & \multicolumn{1}{l|}{R2GenRL \cite{qin-song-2022-reinforced}}          & 0.381                          & 0.232                          & 0.155                          & 0.109                          & 0.287                 & 0.151  &-   & 0.342 & 0.294 & 0.292    \\
                                                       & \multicolumn{1}{l|}{M2KT \cite{yang2023radiology}}                    & 0.386                          & 0.237                          & 0.157                          & 0.111                          & 0.274                  & -  &0.111    &0.420 &0.339 &0.352  \\
                                                       & \multicolumn{1}{l|}{DCL \cite{li2023dynamic}}                          & -                              & -                              & -                              & 0.109                          & 0.284                & 0.150    & -    &0.471 & 0.352  & 0.373\\
                                                       & \multicolumn{1}{l|}{KiUT \cite{huang2023kiut}}                        & 0.393                          & 0.243                          & 0.159                          & 0.113                          & 0.285  
                                                       &0.160 & -& 0.371 & 0.318 & 0.321\\
                                                       & \multicolumn{1}{l|}{PromptMRG \cite{jin2024promptmrg}}                & 0.398                          & -                              & -                              & 0.112                          & 0.268        & 0.157   & -           & 0.501 & 0.509 & 0.476    \\
                                                       & \multicolumn{1}{l|}{DMVF \cite{tang2025dual}}                         & 0.396                          & 0.245                          & 0.162                         & 0.116                         & \textbf{0.289} 
                                                       & 0.154  &0.116&0.451&0.386&0.413\\ 
                                                        & \multicolumn{1}{l|}{DDL-GCN \cite{xu2025deep}}                         & 0.394                          & 0.245                          & 0.165                          & \textbf{0.118 }                        & 0.282 
                                                        & 0.158  &0.118
                                                        &0.488&0.403&0.406
                                                        \\ 
                                                       \cmidrule{2-12}
                                                       & \multicolumn{1}{l|}{\textbf{RG-MVR} (Ours)} & \textbf{0.412}                 & \textbf{0.251}                 & \textbf{0.166}    &\textbf{0.118}       & 0.276
 & \textbf{0.161}         & \textbf{0.142}    & \textbf{0.508} & \textbf{0.518 }& \textbf{0.483}       \\ \midrule
\multicolumn{1}{c|}{\multirow{7}{*}{\textbf{\rotatebox{90}{IU X-ray}}}} & \multicolumn{1}{l|}{R2Gen \cite{chen-etal-2020-generating}}            & 0.289                          & 0.155                          & 0.087                          & 0.052                          & 0.243          & 0.128      & -      &0.151&0.145&0.145     \\
\multicolumn{1}{c|}{}                                  & \multicolumn{1}{l|}{M2KT \cite{yang2023radiology}}                     & 0.371                          & 0.239                          & 0.151                          & 0.078                          & 0.261    & 0.153          &-       &0.153&0.145&0.145      \\
\multicolumn{1}{c|}{}                                  & \multicolumn{1}{l|}{DCL \cite{li2023dynamic}}                          & 0.354                          & 0.230                          & 0.148                          & 0.074                          & 0.267        & 0.152          &-   &0.168&0.167&0.162      \\
\multicolumn{1}{c|}{}                                  & \multicolumn{1}{l|}{RGRG \cite{tanida2023interactive}}                 & 0.266                          & 0.215                          & 0.147                          & 0.063                          & 0.180         & 0.146          &-    &0.183&0.187&0.180    \\
\multicolumn{1}{c|}{}                                  & \multicolumn{1}{l|}{CVT2Dis. \cite{nicolson2023improving}}             & 0.383                          & 0.236                          & 0.157                          & 0.082                          & 0.277      & 0.147            &-    &0.174&0.172&0.168     \\
\multicolumn{1}{c|}{}                                  & \multicolumn{1}{l|}{PromptMRG \cite{jin2024promptmrg}}                 &0.401 &- &- & 0.098 & 0.281 & 0.160 &- &0.213&0.229&0.211\\ 
\multicolumn{1}{c|}{}                                  & \multicolumn{1}{l|}{REVTAF \cite{zhou2025learnable}}                 & 0.420 & 0.249 & 0.159 & 0.107 & 0.309 & \textbf{0.176} &- &\textbf{0.286}&\textbf{0.282}&\textbf{0.273}\\ 
\cmidrule{2-12} 

\multicolumn{1}{c|}{}                                  & \multicolumn{1}{l|}{\textbf{RG-MVR} (Ours)} & \textbf{0.435}                 & \textbf{0.259}                 & \textbf{0.164}                 & \textbf{0.110}                 & \textbf{0.322} & 0.168      & \textbf{0.163}   &0.209&0.212&0.202       \\ \midrule
\end{tabular}}
\label{tab1:sota}
\end{table*}

\begin{table*}[t]
\centering
\caption{Re-evaluation of the PromptMRG baseline and RG-MVR on the MIMIC-CXR test set (3{,}858 samples) using an identical metric pipeline. MIMIC-CXR served as the primary quantitative benchmark; the external DR cohort was reserved for generalization and safety stress testing.}
\label{tab:mimic_primary}
\scalebox{0.68}{
\begin{tabular}{lrrrrrrrrrr}
\toprule
Model & Samples & BLEU-1 & BLEU-2 & BLEU-3 & BLEU-4 & METEOR & ROUGE-L  & Precision & Recall & F1 \\
\midrule
PromptMRG (re-evaluated) & 3{,}858 & 0.387 & 0.230 & 0.147 & 0.100 & 0.147 & 0.261 & 0.506 & 0.466 & 0.455 \\
RG-MVR & 3{,}858 & \textbf{0.412} & \textbf{0.251} & \textbf{0.166} & \textbf{0.118} & \textbf{0.161} & \textbf{0.276}  & \textbf{0.508} & \textbf{0.518} & \textbf{0.483} \\
\bottomrule
\end{tabular}}
\end{table*}

\subsection{Comparison with other methods and ablation study}
\subsubsection{Comparison with other methods}
Following the taxonomy of S2D-ALIGN~\cite{gao2025s2d}, RG-MVR belongs to the from-scratch RRG category. Table~\ref{tab1:sota} compares it with image-captioning baselines (ST~\cite{vinyals2015tell}, Att2In~\cite{rennie2017selfcritical}, AdaAtt~\cite{lu2017knowing}, and TopDown~\cite{anderson2018bottomup}) and representative RRG models, including R2Gen~\cite{chen-etal-2020-generating}, R2GenCMN~\cite{chen-etal-2021-cross-modal}, R2GenRL~\cite{qin-song-2022-reinforced}, M2KT~\cite{yang2023radiology}, DCL~\cite{li2023dynamic}, KiUT~\cite{huang2023kiut}, PromptMRG~\cite{jin2024promptmrg}, DMVF~\cite{tang2025dual}, DDL-GCN~\cite{xu2025deep}, and REVTAF~\cite{zhou2025learnable}.
On MIMIC-CXR, RG-MVR achieved the highest BLEU-1 to BLEU-3 (0.412/0.251/0.166), METEOR (0.161), and CIDEr (0.142), tied with DDL-GCN for the highest BLEU-4 (0.118), and remained competitive on ROUGE-L (0.276 vs.\ 0.289 for DMVF). It also achieved the highest clinical-efficacy (CE) precision, recall, and F1 (0.508/0.518/0.483), exceeding the strongest comparator, PromptMRG (F1, 0.476). Re-evaluating the PromptMRG baseline and RG-MVR with an identical metric pipeline yielded consistent gains across all text and CE metrics (Table~\ref{tab:mimic_primary}). On IU X-ray, RG-MVR achieved the highest BLEU-1 to BLEU-4 and ROUGE-L, but its METEOR (0.168) was lower than that of REVTAF (0.176), and its CE F1 (0.202) remained below those of REVTAF (0.273) and PromptMRG (0.211). 
Representative examples in Fig.~\ref{fig3:env} illustrate differences in generated findings and report content across methods.

\begin{table*}[!t]
\centering
    \caption{
    Single-component ablation on the MIMIC-CXR~\cite{johnson2019mimic} test set. RA-EG and AMER each improve clinical-label recall and F1 over the base model but increase hallucination and false-positive rates when used alone; AMER gives the strongest single-component clinical-label result, and CPL a smaller improvement.
    Combining RA-EG, AMER, and CPL gives the best overall balance, improving clinical-label agreement while reducing hallucination and false-positive rates relative to the base model. FPR, false-positive rate.
    }
    \scalebox{0.63}{
\begin{tabular}{c|ccl|cccccc|ccccc}
\midrule
\multirow{2}{*}{Setting}                                          & \multirow{2}{*}{RA-EG} & \multirow{2}{*}{AMER} & \multirow{2}{*}{CPL} & \multicolumn{6}{c|}{NLG Metrics}                                                                    & \multicolumn{5}{c}{CE Metrics}                         \\ \cmidrule{5-15} 
                                                                  &                       &                       &                       & BLEU-1         & BLEU-2         & BLEU-3         & BLEU-4         & METEOR         & ROUGE-L          & Precision      & Recall         & F1             & Hallucination & FPR  \\ \midrule
Base                                                              &                       &                       &                       & 0.399 & 0.243 & 0.161 & 0.114 & 0.160 & 0.273 & 0.487 & 0.490  & 0.460     & 41.5\%         & 8.6\%             \\ 
(a)                                                               & $\checkmark$          &                       &                       & 0.402 & 0.245 & 0.162 & 0.114 & 0.158 & 0.274 & 0.473 & 0.547 & 0.477 &46.0\% &10.8\%  \\ 
(b)                                                               &                       & $\checkmark$          &                       & 0.405 & 0.247 & 0.164 & 0.115 & 0.158 & 0.274 & 0.480 & \textbf{0.550} & 0.482 &44.7\% &10.5\%  \\ 
(c)                                                               &                       &                       & $\checkmark$ & 0.405 & 0.247 & 0.164 & 0.115 & 0.158 & 0.274 & 0.469 & 0.532 & 0.467 & 43.9\%    & 10.3\%  \\ 
%
Ours & $\checkmark$          & $\checkmark$          & $\checkmark$                      & \textbf{0.412} & \textbf{0.251} & \textbf{0.166} & \textbf{0.118} & \textbf{0.161} & \textbf{0.276} & \textbf{0.508} & 0.518 & \textbf{0.483}           & \textbf{39.1\%}    & \textbf{8.1\%}  \\ \midrule
    \end{tabular}}
    \label{tab3:each component}
\end{table*}

\begin{table*}[t]
    \centering
    \caption{
    Ablation of the learnable precision-weighted multi-task objective on MIMIC-CXR~\cite{johnson2019mimic}. 
    Optimizing either task weight alone provides limited or uneven benefits, and classification-only weighting reduces clinical-label performance. 
    Jointly learning the language-modeling and classification task weights yields the strongest overall language and clinical-label performance while reducing hallucination and false-positive burden.
    }
    \scalebox{0.64}{
    \begin{tabular}{c|cc|cccccc|ccccc} 
        \toprule
        \multirow{2}{*}{Setting}
& \multirow{2}{*}{$\bm{e^{-\eta_{lm}}}$}
& \multirow{2}{*}{$\bm{e^{-\eta_{cls}}}$}
& \multicolumn{6}{c|}{NLG Metrics}
& \multicolumn{5}{c}{CE Metrics} \\
        \cmidrule{4-14}  
        &&& BLEU-1 & BLEU-2 & BLEU-3 & BLEU-4 & METEOR & ROUGE-L & Precision & Recall & F1 & Hallucination   & FPR  \\
        \midrule
        $\mathcal{L}_{LM} + \mathcal{L}_{CLS}$ &&& 0.404 & 0.246 & 0.164 & 0.116 & 0.156 & 0.270 & 0.469 & \textbf{0.527} & 0.467 & 48.1\% & 10.7\% \\
        (a)   & $\checkmark$ && 0.400 & 0.243 & 0.160 & 0.112 & 0.153 & 0.267 & 0.469 & 0.517 & 0.462 &46.7\% &10.4\%  \\
        (b)  & & $\checkmark$& 0.402 & 0.245 & 0.162 & 0.114 & 0.158 & 0.271 & 0.440 & 0.489 & 0.434 &46.8\% &10.8\%  \\
        \midrule
       
        \textbf{Ours} & $\checkmark$          & $\checkmark$                      & \textbf{0.412} & \textbf{0.251} & \textbf{0.166} & \textbf{0.118} & \textbf{0.161} & \textbf{0.276} & \textbf{0.508} & 0.518 & \textbf{0.483}           & \textbf{39.1\%}    & \textbf{8.1\%}    \\
        \bottomrule
    \end{tabular}}
    \label{tab3:loss}
\end{table*}

\subsubsection{Ablation studies}
\noindent\textbf{Effects of individual components.}
%
Table~\ref{tab3:each component} summarizes the component ablation on MIMIC-CXR. Adding RA-EG alone increased recall (0.490$\rightarrow$0.547) and F1 (0.460$\rightarrow$0.477) but also increased the hallucination rate (41.5\%$\rightarrow$46.0\%) and the false-positive rate (FPR; 8.6\%$\rightarrow$10.8\%). AMER alone yielded the highest recall (0.550) and F1 (0.482) among the single-component variants, although both error rates again exceeded those of the base model. CPL alone produced a smaller F1 gain (0.467). Combining all three components yielded the highest F1 (0.483) and the lowest hallucination rate (39.1\%) and FPR (8.1\%) among the evaluated configurations, at the cost of lower recall than RA-EG or AMER alone (0.518 vs.\ 0.547 and 0.550). These results suggest that the components are complementary, with the full model trading a small loss in recall for higher precision and fewer false positives.
\begin{table*}[t]
\centering
\caption{
Composition of the external DR cohort used for generalization and safety stress testing. 
The cohort includes both single-image and multi-image records, as well as cases with focal lesions, pleural or thoracic-cavity abnormalities, postoperative findings, support devices, and aortic calcification or tortuosity. 
These cohort descriptors are descriptive and are not population estimates.}
\label{tab:external_dr_chars}
\begin{tabular}{lrl}
\toprule
Characteristic & Cases & Rate \\
\midrule
Total cases & 22 & 22/22 (100.0\%) \\
Single image per case & 11 & 11/22 (50.0\%) \\
Multi-image per case & 11 & 11/22 (50.0\%) \\
Nodule or focal dense lesion & 7 & 7/22 (31.8\%) \\
Pleural or thoracic cavity abnormality & 21 & 21/22 (95.5\%) \\
Postoperative cases & 4 & 4/22 (18.2\%) \\
Device-related cases & 3 & 3/22 (13.6\%) \\
Aortic calcification or tortuosity & 6 & 6/22 (27.3\%) \\
\bottomrule
\end{tabular}
\end{table*}

\begin{table*}[t]
\centering
\caption{
Automatic metrics on the external DR stress test.
All standard-inference conditions show substantial cross-domain degradation and a high hallucination burden.
Supplying oracle diagnostic states derived from the reference reports markedly improves clinical-label agreement and reduces hallucination, suggesting a bottleneck in disease-state prediction; this condition is a mechanistic upper bound rather than a deployable inference setting.
FPR, false-positive rate; FP, false positive.
}
\label{tab:external_dr_metrics}
\scalebox{0.64}{
\begin{tabular}{lrrrrrrrrrrr}
\toprule
Model & Samples & BLEU-1 & BLEU-4 & ROUGE-L & CIDEr & Precision & Recall & F1 & FPR micro & Hallucination micro & Any FP \\
\midrule
Base & 22 & 0.2379 & 0.0000 & 0.1722 & 0.0305 & 0.0909 & 0.1212 & 0.0985 & 0.1719 & 0.9245 & 0.9091 \\
RG-MVR & 22 & 0.2266 & 0.0150 & 0.1664 & 0.0273 & 0.0833 & 0.0833 & 0.0712 & 0.0737 & 0.8750 & 0.4091 \\
RG-MVR without retrieval & 22 & 0.2134 & 0.0000 & 0.1696 & 0.0107 & 0.0682 & 0.0379 & 0.0485 & 0.0737 & 0.9130 & 0.6818 \\
RG-MVR oracle states & 22 & 0.2630 & 0.0223 & 0.1753 & 0.0397 & 0.5000 & 0.4697 & 0.4621 & 0.0175 & 0.2632 & 0.2273 \\
\bottomrule
\end{tabular}}
\end{table*}
\noindent\textbf{Effect of learnable task weighting.}
Table~\ref{tab3:loss} ablates the task-weighting scheme.
Relative to the fixed-weight objective ($\mathcal{L}_{LM}+\mathcal{L}_{CLS}$), learning only $e^{-\eta_{lm}}$ (a) slightly reduced the hallucination rate (48.1\%$\rightarrow$46.7\%) but marginally lowered NLG scores, whereas learning only $e^{-\eta_{cls}}$ (b) reduced clinical efficacy (F1, 0.467$\rightarrow$0.434).
Learning both weights jointly gave the best overall performance, reducing the hallucination rate from 48.1\% to 39.1\% and the FPR from 10.7\% to 8.1\% while achieving the highest F1 (0.483).
These results indicate that joint task weighting is associated with improved clinical-label agreement and a lower false-positive burden in this setting.
%

\subsection{External DR generalization and safety audit}
\label{Sec:2.4}

To assess generalization beyond public benchmarks, we used an external cohort of 22 chest radiographic examinations from Xinhua Hospital Affiliated to Dalian University, China.
Given its limited size, this cohort was used to assess external generalization and to characterize potential safety-relevant errors rather than to establish model superiority. It comprised 11 single-image and 11 multi-image records; seven cases contained nodular or focal dense lesions, 21 had pleural or thoracic-cavity abnormalities, four were postoperative, three contained support devices, and six showed aortic calcification or tortuosity (Table~\ref{tab:external_dr_chars}). These characteristics were derived from image metadata and reference reports and are reported descriptively.

Automatic metrics showed substantial cross-domain degradation (Table~\ref{tab:external_dr_metrics}). Under standard inference, CheXbert F1 was 0.0985 for Base, 0.0712 for RG-MVR, and 0.0485 for RG-MVR without retrieval. RG-MVR nevertheless produced fewer false-positive labels than Base, with a lower micro-averaged FPR (0.0737 vs.\ 0.1719) and a lower proportion of cases with any false-positive finding (0.4091 vs.\ 0.9091). Supplying oracle diagnostic states raised CheXbert F1 to 0.4621 and lowered the micro-averaged hallucination rate to 0.2632, suggesting that errors in diagnostic-state prediction contribute substantially to the degradation under domain shift. Because oracle states were derived from reference reports, this condition was treated as a mechanistic upper bound rather than as deployable performance.


\begin{table*}[t]
\centering
\caption{
RadGraph semantic robustness on the external DR cohort.  Complete F1, which requires entity, label, and relation agreement, remains low across all inference conditions.
RG-MVR improves strict semantic overlap relative to Base and no-retrieval inference, whereas oracle-state prompting improves CheXbert agreement without a comparable improvement in strict entity--relation overlap.}
\label{tab:external_dr_radgraph}
\begin{tabular}{lrrrr}
\toprule
Model & Samples & Simple F1 & Partial F1 & Complete F1 \\
\midrule
Base & 22 & 0.1115 & 0.0958 & 0.0579 \\
RG-MVR  & 22 & 0.1424 & 0.1299 & 0.0819 \\
RG-MVR without retrieval & 22 & 0.1067 & 0.0917 & 0.0475 \\
RG-MVR oracle states & 22 & 0.1481 & 0.1331 & 0.0798 \\
\bottomrule
\end{tabular}
\end{table*}

\subsection{Uncertainty and clinical semantic robustness}

Case-level bootstrap analysis showed that automatic metrics in the 22-case external cohort were highly uncertain: CheXbert F1 was 0.0985 (95\% CI, 0.0000--0.2273) for Base, 0.0712 (95\% CI, 0.0000--0.1545) for RG-MVR, and 0.0485 (95\% CI, 0.0000--0.1212) for RG-MVR without retrieval, so the lower confidence bound was zero for every standard-inference model. Paired randomization tests showed no statistically significant differences in the main clinical metrics between RG-MVR and its no-retrieval variant, whereas oracle states significantly improved CheXbert F1 and the micro-averaged hallucination rate relative to standard inference. These results support a mechanistic interpretation rather than a claim that retrieval alone determines safety.

As an additional semantic robustness analysis, we computed RadGraph F1 on the same outputs (Table~\ref{tab:external_dr_radgraph}). Complete F1 was low for all models, indicating limited strict entity--relation overlap between generated and reference reports. RG-MVR achieved a complete F1 of 0.0819, compared with 0.0579 for Base, 0.0475 for the no-retrieval variant, and 0.0798 with oracle states. That oracle states markedly improved CheXbert F1 without a comparable gain in RadGraph complete F1 suggests that label-level and entity--relation metrics capture different aspects of report quality.

\begin{table*}[t]
\centering
\caption{
Error modes identified in an earlier exploratory clinician annotation of the external DR cohort. This annotation set is separate from the paired two-physician assessment (Tables~\ref{tab:table10} and~\ref{tab:table11}); the counts were obtained with different procedures and should not be compared directly.
}
\label{tab:external_dr_errors}
\begin{tabular}{lrr}
\toprule
Error mode & Cases & Rate \\
\midrule
Any hallucinated or wrong content & 13 & 13/22 (59.1\%) \\
Unsupported prior-comparison hallucination & 8 & 8/22 (36.4\%) \\
Device-related error & 3 & 3/22 (13.6\%) \\
Postoperative-status error & 2 & 2/22 (9.1\%) \\
Location or laterality error & 3 & 3/22 (13.6\%) \\
Missed major finding & 15 & 15/22 (68.2\%) \\
Any major issue & 21 & 21/22 (95.5\%) \\
\bottomrule
\end{tabular}
\end{table*}
\begin{table*}[t]
\centering
\caption{Descriptive results of the two-physician review of the 22-case external DR cohort. Counts represent non-empty annotation fields or keyword mentions, not adjudicated clinical error rates.}
\label{tab:table10}
\small
\setlength{\tabcolsep}{5pt}
\begin{tabular}{lrrr}
\toprule
Review item & Base (n=22) & RG-MVR (n=22) & Exact reviewer agreement \\
\midrule
\multicolumn{4}{l}{\textit{Binary field-presence / keyword screening, n/N (\%)}} \\
``Right'' field present & 15/22 (68.2) & 16/22 (72.7) & 22/22 (100) \\
``Wrong'' field present & 22/22 (100) & 22/22 (100) & 22/22 (100) \\
``Missed'' field present & 22/22 (100) & 22/22 (100) & 22/22 (100) \\
Prior-comparison mention & 22/22 (100) & 9/22 (40.9) & 22/22 (100) \\
Device mention & 13/22 (59.1) & 7/22 (31.8) & 22/22 (100) \\
Postoperative mention & 4/22 (18.2) & 2/22 (9.1) & 22/22 (100) \\
Pneumothorax/effusion mention & 22/22 (100) & 22/22 (100) & 22/22 (100) \\
Parenchymal-finding mention & 21/22 (95.5) & 10/22 (45.5) & 22/22 (100) \\
Location/laterality mention & 22/22 (100) & 21/22 (95.5) & 22/22 (100) \\
\midrule
\multicolumn{4}{l}{\textit{Normalized exact agreement on free-text annotations, n/N (\%)}} \\
``Right'' text & 22/22 (100) & 21/22 (95.5) & -- \\
``Wrong'' text & 11/22 (50.0) & 12/22 (54.5) & -- \\
``Missed'' text & 21/22 (95.5) & 21/22 (95.5) & -- \\
\bottomrule
\end{tabular}
\vspace{2pt}
\begin{minipage}{0.98\textwidth}\footnotesize Binary exact agreement compares the two reviewers' case-level present/absent coding. Cohen's $\kappa$ was 1.0 for non-constant binary indicators for which it was estimable; it is undefined for constant indicators. Text agreement is exact agreement after normalization and is shown separately from binary coding agreement. Keyword counts are based on mentions in ``wrong''/``missed'' text and do not establish whether a finding was erroneous.\end{minipage}
\end{table*}

\begin{table*}[t]
\centering
\caption{Two-physician assessment of report-level error endpoints in the external DR cohort. Values are the number of cases with each endpoint (percentage); the difference is RG-MVR minus Base in percentage points.}
\label{tab:table11}
\small
\setlength{\tabcolsep}{7pt}
\begin{tabular}{lrrrr}
\toprule
Endpoint & Base (n=22) & RG-MVR (n=22) & Difference (pp) & $P$ value \\
\midrule
Any incorrect or unsupported statement & 22 (100.0\%) & 20 (90.9\%) & $-9.1$ & 0.500 \\
Unsupported prior/temporal comparison & 20 (90.9\%) & 9 (40.9\%) & $-50.0$ & 0.001 \\
Unsupported device/procedure-related statement & 15 (68.2\%) & 7 (31.8\%) & $-36.4$ & 0.021 \\
\bottomrule
\end{tabular}
\vspace{2pt}
\begin{minipage}{0.98\textwidth}\footnotesize $P$ values were obtained from two-sided exact McNemar tests with the case as the paired unit and are not adjusted for multiple comparisons.\end{minipage}
\end{table*}

\subsection{Clinician assessment of report factuality}
To assess report factuality beyond automatic metrics, we analyzed clinician annotations of the 22 external DR examinations described in Section~\ref{Sec:2.4}. Two annotation sets were available: an earlier exploratory annotation of reporting errors and a separate paired two-physician assessment of reports generated by Base and RG-MVR. Because they differed in annotation procedures and error definitions, the two sets were analyzed separately and were not pooled.

The earlier exploratory annotations identified incorrect or unsupported content in 13 cases (59.1\%), missed major findings in 15 (68.2\%), and at least one major issue in 21 (95.5\%); unsupported prior comparisons, device-related errors, postoperative-status errors, and location or laterality errors were also recorded (Table~\ref{tab:external_dr_errors}). These results provide a descriptive overview of clinically relevant error patterns.

In the two-physician assessment, report-level endpoints were compared between paired Base and RG-MVR reports (Table~\ref{tab:table11}). At least one incorrect or unsupported statement was present in 22 Base reports (100.0\%) and 20 RG-MVR reports (90.9\%), a difference that was not statistically significant ($P=0.500$). RG-MVR reports, however, less often contained unsupported prior or temporal comparisons (9 vs.\ 20 cases; 40.9\% vs.\ 90.9\%; $P=0.001$) and unsupported device- or procedure-related statements (7 vs.\ 15 cases; 31.8\% vs.\ 68.2\%; $P=0.021$). Thus, RG-MVR reduced specific categories of unsupported content, whereas report-level errors overall remained frequent.

The reviewers also recorded free-text ``right'', ``wrong'', and ``missed'' annotations for each case (Table~\ref{tab:table10}). The ``right'' field contained text for 15 Base and 16 RG-MVR cases, whereas the ``wrong'' and ``missed'' fields contained text for all cases under both conditions. Keyword screening of the ``wrong'' and ``missed'' annotations identified fewer RG-MVR than Base cases with mentions of prior comparison (9 vs.\ 22), devices (7 vs.\ 13), postoperative findings (2 vs.\ 4), and parenchymal findings (10 vs.\ 21); pneumothorax/effusion mentions occurred in all cases under both conditions, and location/laterality mentions in 21 vs.\ 22 cases. Because a keyword can also be triggered by negated or non-error text, these counts reflect mentions rather than error rates. Binary field-presence and keyword indicators agreed exactly between the two reviewers, whereas normalized exact agreement on the free-text annotations was lower (Base: 22/22, 11/22, and 21/22 for ``right'', ``wrong'', and ``missed''; RG-MVR: 21/22, 12/22, and 21/22).

These clinician assessments should be regarded as exploratory. Model identities in the annotation records were assigned by matching the annotated report text with the generated reports, so the exact files presented to the reviewers could not be independently confirmed. The assessment was not designed as a formal blinded reader study with randomized report presentation and adjudication of disagreements by a senior radiologist; it identifies safety-relevant failure modes but does not replace such a study. Before clinical claims can be made, a reader study with anonymized and randomized reports, at least two blinded radiologists, predefined adjudication rules, and ratings of major errors, missed and unsupported findings, prior-comparison, device, postoperative, and laterality errors, clinical acceptability, and suitability as a draft report will be required.

\section{Discussion}

This study presents RG-MVR, a retrieval-grounded framework that integrates evidence refinement, adaptive multi-view fusion, and diagnostic-state prompting for chest radiograph report generation. On MIMIC-CXR, RG-MVR achieved the highest clinical-label agreement and competitive or superior language-generation performance among the compared methods. The ablation results distinguish the effects of individual components from those of their combination: RA-EG and AMER each increased clinical-label recall but also increased the automatically measured hallucination and false-positive rates, whereas the complete framework achieved the highest clinical-label F1 and the lowest values of both error measures among the evaluated configurations. This pattern suggests that retrieval refinement, evidence fusion, and diagnostic prompting are more effective in combination than in isolation under the evaluated conditions.

The methodological contribution of RG-MVR lies in explicitly separating evidence refinement, evidence integration, and diagnostic-state representation. RA-EG combines dense memory representations with sparse retrieved evidence and uses visual gating to modulate the contribution of retrieved information according to the current image. AMER weights visual, memory, and retrieved representations case by case, allowing their relative contributions to vary across examinations. CPL separates finding mention from diagnostic polarity, providing structured diagnostic information to the decoder. Together, these components target distinct sources of uncertainty in retrieval-augmented generation. The ablation results are consistent with complementary effects; however, whether the learned fusion weights reflect clinically reliable evidence selection requires direct case-level examination.

The cross-dataset results indicate that gains in linguistic quality do not necessarily translate into comparable gains in diagnostic-label agreement. On IU X-ray, RG-MVR achieved favorable language-generation scores, but its clinical-label F1 remained below those of REVTAF and PromptMRG. In the external cohort from Xinhua Hospital Affiliated to Dalian University, the CheXbert F1 of RG-MVR fell to 0.0712, below that of Base (0.0985), although the bootstrap confidence intervals of all standard-inference models overlapped widely. Removing retrieval did not resolve this degradation, whereas supplying reference-derived diagnostic states markedly improved label-level performance. The latter result points to diagnostic-state prediction as a likely contributor to cross-domain errors, although it does not isolate this factor from differences in imaging characteristics, reporting conventions, or retrieved evidence. Moreover, oracle-state prompting relies on information unavailable at inference time and cannot be interpreted as deployable performance.

Clinician assessments captured errors that automatic metrics do not adequately reflect. Compared with Base, RG-MVR produced fewer reports containing unsupported prior or temporal comparisons (9 vs.\ 20) and unsupported device- or procedure-related statements (7 vs.\ 15). However, the proportion of examinations with any incorrect or unsupported statement remained high, with no statistically significant difference between models; the reduction in specific categories of unsupported content therefore does not establish an overall improvement in clinical reliability. This distinction is particularly relevant to radiology reporting, in which descriptions of interval change, procedural history, or device status may depend on clinical information or prior examinations that are unavailable to the model. CheXbert-based evaluation measures agreement within a predefined label set, whereas RadGraph captures aspects of entity--relation correspondence; neither determines whether every generated statement is supported by the available evidence. Physician assessment therefore complements, rather than replaces, automatic evaluation.

This study has several limitations. The external cohort contains only 22 examinations from a single institution, limiting the precision and generalizability of the observed differences. The physician assessments are exploratory, and the provenance and blinding procedures of the archived annotations have not been fully verified. Clinical-label evaluation is restricted to a predefined set of findings and does not capture anatomical localization, severity, or all clinically important omissions. The retrieval memory is constructed from MIMIC-CXR, and the effects of demographic and institutional differences have not been systematically evaluated. Although AMER uses adaptive fusion weights, their relationship to the accuracy of the retrieved evidence remains unclear. Finally, the absence of prior examinations limits the assessment of temporal descriptions. Larger multi-institutional cohorts, standardized independent clinical review, and case-level analyses of retrieved evidence are needed to determine whether the benchmark improvements translate into fewer clinically important reporting errors.

In conclusion, RG-MVR improves clinical-label agreement on the primary MIMIC-CXR benchmark by combining evidence-aware retrieval, adaptive multi-view fusion, and structured diagnostic prompting, and in an exploratory external cohort it reduced specific categories of unsupported statements. The marked degradation under domain shift and the persistently high rate of report-level errors indicate that such systems should currently be regarded as draft-generation aids under radiologist oversight rather than as autonomous reporting tools, and that institution-specific validation will be essential before clinical use.

\section{Methods}
%
\subsection{Network architecture}
Figure~\ref{fig1:all} gives an overview of RG-MVR, a retrieval-augmented encoder--decoder whose design loosely mirrors the observe--recall--reason--write sequence of radiological reporting.
An ImageNet-pretrained ResNet-101~\cite{HZR16} visual encoder produces patch tokens $\mathbf{P}$ and a pooled feature $\bar{\mathbf{v}}$, a BERT-base~\cite{DCL19} decoder generates the report, and a CLIP-encoded~\cite{radford2021clip} report memory constructed from the training set provides auxiliary evidence~\cite{jin2024promptmrg}.
RA-EG constructs a dense memory summary and sparse top-$k$ evidence and applies visual-gated denoising to suppress retrieval noise.
AMER adaptively fuses visual cues, dense priors, and refined sparse evidence into a diagnosis-oriented representation $\mathbf{f}_{\text{final}}$.
CPL converts predicted disease states into discrete prompt tokens for the decoder, and the model is trained with the learnable precision-weighted multi-task objective described below, which combines language modeling and diagnostic supervision.

\subsection{(a) Retrieval-Augmented Evidence Grounding (RA-EG)}  
As shown in Fig.~\ref{fig1:all}(a), RA-EG is designed to exploit the CLIP-encoded report memory used in memory-based prompting~\cite{jin2024promptmrg} more robustly.
It jointly models (i) a dense memory-summarized prior and (ii) a sparse top-$k$ evidence vector, the latter refined by a visual-gated denoising mechanism that suppresses noisy candidates.

Let $\mathbf{M}\in\mathbb{R}^{B\times N\times C}$ denote the retrieved memory features for a mini-batch, where $N$ is the number of retrieved candidates and $C=512$.
The visual encoder outputs a pooled feature $\bar{\mathbf{v}}\in\mathbb{R}^{B\times D}$ and patch-level features $\mathbf{P}\in\mathbb{R}^{B\times L\times D}$ with $D=2048$.
For notational simplicity, we omit the batch dimension $B$ in the following.

\textbf{Dense memory summarization.}
A Transformer-based memory module $\mathrm{MemTrans}(\cdot)$ summarizes global priors from $\mathbf{M}$ using a visual query constructed from $\bar{\mathbf{v}}$:
\begin{equation}
\mathbf{m}=\mathrm{MemTrans}\bigl(\mathbf{M},\,\mathbf{q}_{m}\bigr),\qquad
\mathbf{q}_{m}=\phi_{q}(\bar{\mathbf{v}}),
\end{equation}
where $\phi_{q}(\cdot)$ is a learnable projection to the memory-query space.

\textbf{Sparse top-$k$ evidence aggregation.}
In parallel, we construct a retrieval query by combining pooled and patch-level cues:
\begin{equation}
\mathbf{q}=\mathrm{Norm}\Bigl(\phi_{q}(\bar{\mathbf{v}})+\phi_{q}\bigl(\mathrm{Mean}_{L}(\mathbf{P})\bigr)\Bigr),
\end{equation}
where $\mathrm{Mean}_{L}(\cdot)$ denotes mean pooling over the patch/token dimension and $\mathrm{Norm}(\cdot)$ is $\ell_2$ normalization.
We then project each memory entry into key/value embeddings:
\begin{equation}
\mathbf{k}_i=\mathrm{Norm}\bigl(\phi_{\mathrm{key}}(\mathbf{M}_i)\bigr),\qquad
\mathbf{u}_i=\phi_{\mathrm{val}}(\mathbf{M}_i),
\end{equation}
where $\phi_{\mathrm{key}}(\cdot)$ and $\phi_{\mathrm{val}}(\cdot)$ are lightweight projections.
Cosine similarities are computed as $s_i=\mathbf{k}_i^\top \mathbf{q}$, 
and we select the set of indices $\mathcal{I}_{k}$ corresponding to the $k$ most relevant memory entries:
\begin{equation}
\mathcal{I}_{k}=\mathrm{TopK}(\{s_i\}_{i=1}^{N},\,k),
\end{equation}
where $k$ denotes the number of selected entries.
The sparse evidence is aggregated by temperature-scaled attention over the selected entries, in the spirit of top-$k$ sparse attention~\cite{gupta2021memory,yuan2025native}:
\begin{equation}
w_i=\mathrm{Softmax}\!\left(\frac{s_i}{\tau}\right),\ i\in\mathcal{I}_{k},\qquad
\mathbf{r}=\sum_{i\in\mathcal{I}_{k}} w_i\,\mathbf{u}_i,
\end{equation}
where $\tau>0$ is a temperature hyper-parameter.

\textbf{Visual-gated denoising and grounding.}
To mitigate noisy retrieval, we refine the sparse evidence with a feature-wise visual gate:
\begin{equation}
\tilde{\mathbf{v}}=\phi_{g}(\bar{\mathbf{v}}),\qquad
\mathbf{g}=\sigma\bigl(\mathbf{W}_g[\tilde{\mathbf{v}};\mathbf{r}]\bigr),\qquad
\mathbf{e}=\mathbf{g}\odot\mathbf{r}+(1-\mathbf{g})\odot\tilde{\mathbf{v}},
\end{equation}
where $\mathbf{W}_g$ is a learnable weight matrix, $\phi_{g}(\cdot)$ is a learnable projection, $[\cdot;\cdot]$ denotes the concatenation operation, $\sigma(\cdot)$ denotes the sigmoid function, and $\odot$ denotes the element-wise (Hadamard) product.
The resulting evidence vector $\mathbf{e}$ is thus explicitly conditioned on the image feature.

Overall, RA-EG outputs a dense memory prior $\mathbf{m}$ and a refined sparse evidence vector $\mathbf{e}$, providing complementary global context and high-confidence evidence for subsequent multi-view reasoning and diagnosis-conditioned generation.

\subsection{(b) Adaptive Multi-view Evidential Reasoning (AMER)}
As shown in Fig.~\ref{fig1:all}(b), we propose Adaptive Multi-view Evidential Reasoning (AMER) to integrate the hybrid evidence grounded by RA-EG, including the pooled visual feature $\bar{\mathbf{v}}$, the dense memory prior $\mathbf{m}$, and the refined sparse evidence $\mathbf{e}$.
AMER performs case-adaptive weighting over these three views and further applies residual refinement to strengthen robustness.
%
Since $\mathbf{m}$ and $\mathbf{e}$ lie in a 512-dimensional space, we project $\bar{\mathbf{v}}$ to the same dimensionality, denoted as $\hat{\mathbf{v}}$.
%
We concatenate the three streams as $\mathbf{z}=[\hat{\mathbf{v}};\mathbf{m};\mathbf{e}]$ and compute case-adaptive importance weights:
\begin{equation}
\boldsymbol{\alpha}=\mathrm{Softmax}\!\Big(\mathbf{W}_{gate}\tanh(\mathbf{W}_{proj}\mathbf{z})\Big),
\qquad \boldsymbol{\alpha}=[\alpha_{1},\alpha_{2},\alpha_{3}]\in\mathbb{R}^{3},
\end{equation}
where $\mathbf{W}_{gate}$ and $\mathbf{W}_{proj}$ denote learnable parameters for the gating mechanism, $\alpha_{1}$, $\alpha_{2}$, and $\alpha_{3}$ denote the importance of $\hat{\mathbf{v}}$, $\mathbf{m}$, and $\mathbf{e}$, respectively.
The fused multi-view evidence is obtained by a weighted combination:
\begin{equation}
\mathbf{f}=\alpha_{1}\hat{\mathbf{v}}+\alpha_{2}\mathbf{m}+\alpha_{3}\mathbf{e}.
\end{equation}

To further refine the fused evidence, we incorporate a residual projection from the same concatenated representation:
\begin{equation}
\tilde{\mathbf{f}}=\mathbf{f}+\lambda\cdot \mathbf{W}_{res}\mathbf{z},
\end{equation}
where $\mathbf{W}_{res}$ is a learnable projection and $\lambda$ is a scalar scaling factor controlling residual strength.
Finally, AMER outputs the diagnosis-oriented representation by concatenating the projected visual embedding $\hat{\mathbf{v}}$, the refined fused evidence $\tilde{\mathbf{f}}$, and the refined sparse evidence $\mathbf{e}$:
\begin{equation}
\mathbf{f}_{final} = [\hat{\mathbf{v}}; \tilde{\mathbf{f}}; \mathbf{e}].
\end{equation}

Overall, AMER performs adaptive multi-view evidential reasoning to balance visual grounding with dense priors and sparse retrieved evidence, producing a robust representation for subsequent clinical prompting and report generation.

\subsection{(c) Clinical Prompt Learner (CPL)}
To bridge visual perception and clinically aligned report drafting, we propose a Clinical Prompt Learner (CPL).
As shown in Fig.~\ref{fig1:all}(c), following diagnosis-conditioned prompting~\cite{jin2024promptmrg}, we map per-disease state predictions to discrete prompt tokens and prepend them to the decoder input.
Unlike a flat four-way classifier, CPL uses a decoupled hierarchical design that separates whether a finding is mentioned from its polarity, and couples the two heads softly to avoid the error propagation of hard two-stage routing.

\textbf{Decoupled hierarchical diagnosis.}
Given the fused representation $\mathbf{f}_{final}$, a shared MLP extracts a latent clinical feature $\mathbf{h}$.
We then branch $\mathbf{h}$ into two specialized heads:
(i) a \emph{Presence Head} that predicts a scalar logit $z^{(c)}_{\mathrm{pres}}\in\mathbb{R}$ for each disease $c$, indicating evidence of existence; and
(ii) a \emph{Polarity Head} that outputs a polarity vector $\mathbf{z}^{(c)}_{\mathrm{pol}}\in\mathbb{R}^{3}$ over the non-blank set $\mathcal{S}=\{\texttt{POS},\texttt{NEG},\texttt{UNC}\}$.
We define the diagnosis state set as $\mathcal{Y}=\{\texttt{BLA},\texttt{POS},\texttt{NEG},\texttt{UNC}\}$, where
\texttt{BLA} denotes a blank (not mentioned) finding, \texttt{POS} a positive mention, \texttt{NEG} an explicitly negated mention, and \texttt{UNC} an uncertain mention.

\textbf{Soft-coupled logit composition.}
Let $\mathbf{z}^{(c)}_{\mathrm{pol}}=[z^{(c)}_{\texttt{POS}}, z^{(c)}_{\texttt{NEG}}, z^{(c)}_{\texttt{UNC}}]$.
Instead of hard masking, we compose 4-way logits via soft coupling:
\begin{equation}
y^{(c)}_{\texttt{BLA}}=-z^{(c)}_{\mathrm{pres}},\qquad
y^{(c)}_{s}=z^{(c)}_{s}+z^{(c)}_{\mathrm{pres}},\ \forall s\in\mathcal{S}.
\end{equation}

Here, $z^{(c)}_{\mathrm{pres}}$ acts as a global bias: a higher presence score suppresses \texttt{BLA} while boosting all non-blank polarities.
The margin between any clinical state $s$ and \texttt{BLA} becomes
\begin{equation}
y^{(c)}_{s}-y^{(c)}_{\texttt{BLA}} = z^{(c)}_{s} + 2z^{(c)}_{\mathrm{pres}},\qquad \forall s\in\mathcal{S},
\end{equation}
which helps reduce confusion between unmentioned findings (\texttt{BLA}) and uncertain mentions (\texttt{UNC}) under long-tailed distributions.
The final state is predicted by a single 4-way softmax/argmax over $\mathcal{Y}$.

\textbf{Discrete prompt construction.}
For each disease $c$, we obtain $\hat{y}^{(c)}=\arg\max_{y\in\mathcal{Y}} y^{(c)}_{y}$ and map it to a discrete token in $\{[\texttt{BLA}],[\texttt{POS}],[\texttt{NEG}],[\texttt{UNC}]\}$.
Concatenating tokens across diseases forms a prompt prefix, which is prepended to the decoder input to guide report generation.

\textbf{Clinical-aware logit adjustment (training only).}
To alleviate severe class imbalance, during training we apply a POS-only prior adjustment:
\begin{equation}
y^{(c)}_{\texttt{POS}} \leftarrow y^{(c)}_{\texttt{POS}} + \log s^{(c)}_{\text{base}},
\end{equation}
where $s^{(c)}_{\text{base}}$ is a pre-computed prior statistic for disease $c$.
This term is only used for the diagnosis loss and is discarded during inference to keep the prompt construction unbiased.

\subsection{Training objective}
Report generation and disease classification are optimized jointly.
Report generation uses a teacher-forced language-modeling loss $\mathcal{L}_{LM}$ (token-level cross-entropy, with prompt prefixes masked), and diagnosis uses the focal loss~\cite{lin2017focal} $\mathcal{L}_{CLS}$ to address class imbalance.
To avoid a fixed trade-off between the two tasks, we introduce learnable scalar parameters $\eta_{lm}$ and $\eta_{cls}$ that act as exponential precision-like task weights~\cite{kendall2018multi}:

\begin{equation}
\mathcal{L}_{total} = e^{-\eta_{lm}} \mathcal{L}_{LM} + e^{-\eta_{cls}} \mathcal{L}_{CLS}.
\end{equation}

Both parameters are optimized jointly with the model parameters. Unlike the full homoscedastic-uncertainty formulation~\cite{kendall2018multi}, the additive log-variance regularization terms are omitted; the objective is therefore a practical precision-weighted loss rather than a calibrated uncertainty model. Training remained stable under the evaluated settings.

\subsection{Model evaluation}
Evaluation followed a tiered design. MIMIC-CXR served as the primary benchmark for model comparison and ablation, and IU X-ray was used only as an external cross-dataset validation set, with all 4{,}168 available records used for testing.
The local DR cohort was used only as an external generalization and safety stress-test cohort.

For MIMIC-CXR, all metrics were computed on the 3{,}858 test samples from the reports generated by RG-MVR and by the PromptMRG baseline, using the same evaluation pipeline. Language-generation quality was assessed with BLEU-1 to BLEU-4~\cite{PRW02}, METEOR~\cite{DeL11}, ROUGE-L~\cite{Lin04}, and, where available, CIDEr~\cite{vedantam2015cider}. Clinical-label agreement was assessed with precision, recall, and F1 computed from labels extracted by CheXbert~\cite{SJR20}.

For the external DR cohort, we evaluated Base, RG-MVR, RG-MVR without retrieval, and RG-MVR with oracle diagnostic states on the same 22 cases. In addition to text and CheXbert metrics, we report the micro-averaged false-positive rate, the micro-averaged hallucination rate, and the case-level proportion of reports with any false-positive finding. CheXbert labels were binarized with a positive-only rule: positive labels were mapped to 1, and blank (not mentioned), negative, and uncertain labels to 0.

Semantic robustness on the external DR cohort was further assessed with RadGraph F1~\cite{jain2021radgraph}. We report simple, partial, and complete F1, with complete F1, which requires agreement in entities, labels, and relations, as the strictest endpoint. Composite metrics such as RadCliQ and GREEN could not be computed reproducibly in our environment and are therefore not reported.

\subsection{Clinician assessment}
Two physicians assessed the Base and RG-MVR reports for the same 22 external DR cases. Three binary report-level endpoints were evaluated: any incorrect or unsupported statement, an unsupported prior or temporal comparison, and an unsupported device- or procedure-related statement. Proportions were calculated with the 22 cases as the denominator, and differences were expressed in percentage points (RG-MVR minus Base). Paired comparisons used nominal two-sided exact McNemar tests with the case as the paired unit. 

\subsection{Statistical analysis}
For the external DR cohort, Wilson score intervals were used for binary clinician-derived proportions and model-level any-false-positive rates. Case-level bootstrap confidence intervals were computed from 1{,}000 replicates (random seed 20260824), with the case as the resampling unit and text and CheXbert metrics recomputed on each resample. Paired model comparisons used case-level paired randomization tests with 2{,}000 permutations (seed 20260824), with Benjamini--Hochberg correction across tests.
Case-level any-false-positive indicators and clinician-assessed error endpoints were compared using two-sided exact McNemar tests.

\newpage
\section*{Declaration statements} 
\subsection*{Competing interests}
The authors declare no competing interests. 
\subsection*{Ethics approval and consent to participate}
This study used the publicly available, de-identified MIMIC-CXR~\cite{johnson2019mimic} and IU X-ray~\cite{DKR16} datasets for the primary benchmark experiments. A local de-identified DR cohort was analyzed only as an external generalization and safety stress-test cohort. 
The MIMIC-CXR dataset was collected under a protocol approved by the Institutional Review Board of Beth Israel Deaconess Medical Center, with a waiver of informed consent, and was released in a fully de-identified form in compliance with the Health Insurance Portability and Accountability Act (HIPAA) Safe Harbor provision. Access to MIMIC-CXR was obtained by the authors through PhysioNet under a data use agreement requiring completion of CITI human subjects research training. 
The IU X-ray dataset is a publicly released, de-identified collection made available by the original curators for research use. The local DR cohort is not publicly released because of institutional governance and privacy restrictions.
\subsection*{Data availability} 
MIMIC-CXR is available via PhysioNet (\url{https://physionet.org/content/mimic-cxr/}) under credentialed access and a data use agreement. IU X-ray is publicly available through the Indiana University chest X-ray collection. The local DR images and reports are not publicly available because of institutional governance and privacy restrictions; aggregate evaluation results are reported in this article to support the reproducibility of the external stress-test analyses.

\subsection*{Code availability} 
Code and analysis scripts for MIMIC-CXR evaluation, external DR bootstrap confidence intervals, paired randomization tests, and RadGraph robustness analysis will be made available in a project-specific repository upon publication.

\subsection*{Acknowledgements} 

This work was supported in part by the National Natural Science Foundation of China (No. 62502064), the Joint Plan of the Liaoning Province Science and Technology Plan (Nos. 2025JH2/101800417 and 2025JH2/101800422), the Scientific Research Project of the Liaoning Provincial Department of Education (No. LJ222511258003), the Interdisciplinary Project of Dalian University (No. DLUXK-2025-QN-020), and the 111 Center (No. D23006).


\newpage
\bibliography{sample} 

\begin{thebibliography}{10}
\urlstyle{rm}
\expandafter\ifx\csname url\endcsname\relax
  \def\url#1{\texttt{#1}}\fi
\expandafter\ifx\csname urlprefix\endcsname\relax\def\urlprefix{URL }\fi
\expandafter\ifx\csname doiprefix\endcsname\relax\def\doiprefix{DOI: }\fi
\providecommand{\bibinfo}[2]{#2}
\providecommand{\eprint}[2][]{\url{#2}}

\bibitem{PengFSLYZWZ24}
\bibinfo{author}{Peng, P.} \emph{et~al.}
\newblock \bibinfo{journal}{\bibinfo{title}{Eye gaze guided cross-modal alignment network for radiology report generation}}.
\newblock {\emph{\JournalTitle{{IEEE} J. Biomed. Health Informatics}}} \textbf{\bibinfo{volume}{28}}, \bibinfo{pages}{7406--7419} (\bibinfo{year}{2024}).

\bibitem{2026WZY}
\bibinfo{author}{Wang, Z.} \emph{et~al.}
\newblock \bibinfo{journal}{\bibinfo{title}{Parallel multi-attention and gated fusion for visual question localized answering in surgical scenes}}.
\newblock {\emph{\JournalTitle{{IEEE} J. Biomed. Health Informatics}}} \textbf{\bibinfo{volume}{30}}, \bibinfo{pages}{5748--5760} (\bibinfo{year}{2026}).

\bibitem{2024cfh}
\bibinfo{author}{Cai, F.}, \bibinfo{author}{He, J.}, \bibinfo{author}{Liu, Y.} \& \bibinfo{author}{Zhang, H.}
\newblock \bibinfo{journal}{\bibinfo{title}{Bcslinker: automatic method for constructing a knowledge graph of venous thromboembolism based on joint learning}}.
\newblock {\emph{\JournalTitle{Front. Med.}}} \textbf{\bibinfo{volume}{Volume 11 - 2024}} (\bibinfo{year}{2024}).

\bibitem{huang2023kiut}
\bibinfo{author}{Huang, Z.}, \bibinfo{author}{Zhang, X.} \& \bibinfo{author}{Zhang, S.}
\newblock \bibinfo{title}{Kiut: Knowledge-injected u-transformer for radiology report generation}.
\newblock In \emph{\bibinfo{booktitle}{CVPR}}, \bibinfo{pages}{19809--19818} (\bibinfo{year}{2023}).

\bibitem{hou2025rrg}
\bibinfo{author}{Hou, X.}, \bibinfo{author}{Li, X.}, \bibinfo{author}{Lu, M.}, \bibinfo{author}{Wang, S.} \& \bibinfo{author}{Zhang, Y.}
\newblock \bibinfo{title}{Rrg-mamba: Efficient radiology report generation with state space model}.
\newblock In \emph{\bibinfo{booktitle}{IJCAI}}, \bibinfo{pages}{7410--7418} (\bibinfo{year}{2025}).

\bibitem{lang2025dacg}
\bibinfo{author}{Lang, W.}, \bibinfo{author}{Liu, Z.} \& \bibinfo{author}{Zhang, Y.}
\newblock \bibinfo{journal}{\bibinfo{title}{Dacg: Dual attention and context guidance model for radiology report generation}}.
\newblock {\emph{\JournalTitle{Med. Image Anal.}}} \textbf{\bibinfo{volume}{99}}, \bibinfo{pages}{103377} (\bibinfo{year}{2025}).

\bibitem{jin2024promptmrg}
\bibinfo{author}{Jin, H.}, \bibinfo{author}{Che, H.}, \bibinfo{author}{Lin, Y.} \& \bibinfo{author}{Chen, H.}
\newblock \bibinfo{title}{Promptmrg: Diagnosis-driven prompts for medical report generation}.
\newblock In \emph{\bibinfo{booktitle}{AAAI}}, vol.~\bibinfo{volume}{38}, \bibinfo{pages}{2607--2615} (\bibinfo{year}{2024}).

\bibitem{zhou2025learnable}
\bibinfo{author}{Zhou, Q.} \emph{et~al.}
\newblock \bibinfo{title}{Learnable retrieval enhanced visual-text alignment and fusion for radiology report generation}.
\newblock In \emph{\bibinfo{booktitle}{ICCV}}, \bibinfo{pages}{22529--22538} (\bibinfo{year}{2025}).

\bibitem{xu2025deep}
\bibinfo{author}{Xu, L.} \emph{et~al.}
\newblock \bibinfo{journal}{\bibinfo{title}{Deep disease label-guided graph convolutional network for medical report generation}}.
\newblock {\emph{\JournalTitle{ACM Trans. Knowl. Discov. Data}}} \textbf{\bibinfo{volume}{19}}, \bibinfo{pages}{1--23} (\bibinfo{year}{2025}).

\bibitem{johnson2019mimic}
\bibinfo{author}{Johnson, A.~E.} \emph{et~al.}
\newblock \bibinfo{journal}{\bibinfo{title}{Mimic-cxr, a de-identified publicly available database of chest radiographs with free-text reports}}.
\newblock {\emph{\JournalTitle{Sci. Data}}} \textbf{\bibinfo{volume}{6}}, \bibinfo{pages}{317} (\bibinfo{year}{2019}).

\bibitem{DKR16}
\bibinfo{author}{Demner-Fushman, D.} \emph{et~al.}
\newblock \bibinfo{journal}{\bibinfo{title}{Preparing a collection of radiology examinations for distribution and retrieval}}.
\newblock {\emph{\JournalTitle{J. Am. Med. Inf. Assoc.}}} \textbf{\bibinfo{volume}{23}}, \bibinfo{pages}{304--310} (\bibinfo{year}{2016}).

\bibitem{chen-etal-2020-generating}
\bibinfo{author}{Chen, Z.}, \bibinfo{author}{Song, Y.}, \bibinfo{author}{Chang, T.-H.} \& \bibinfo{author}{Wan, X.}
\newblock \bibinfo{title}{Generating radiology reports via memory-driven transformer}.
\newblock In \emph{\bibinfo{booktitle}{EMNLP}}, \bibinfo{pages}{1439--1449} (\bibinfo{year}{2020}).

\bibitem{seyyed2021underdiagnosis}
\bibinfo{author}{Seyyed-Kalantari, L.}, \bibinfo{author}{Zhang, H.}, \bibinfo{author}{McDermott, M. B.~A.}, \bibinfo{author}{Chen, I.~Y.} \& \bibinfo{author}{Ghassemi, M.}
\newblock \bibinfo{journal}{\bibinfo{title}{Underdiagnosis bias of artificial intelligence algorithms applied to chest radiographs in under-served patient populations}}.
\newblock {\emph{\JournalTitle{Nature Medicine}}} \textbf{\bibinfo{volume}{27}}, \bibinfo{pages}{2176--2182} (\bibinfo{year}{2021}).

\bibitem{vinyals2015tell}
\bibinfo{author}{Vinyals, O.}, \bibinfo{author}{Toshev, A.}, \bibinfo{author}{Bengio, S.} \& \bibinfo{author}{Erhan, D.}
\newblock \bibinfo{title}{Show and tell: {A} neural image caption generator}.
\newblock In \emph{\bibinfo{booktitle}{CVPR}}, \bibinfo{pages}{3156--3164} (\bibinfo{year}{2015}).

\bibitem{rennie2017selfcritical}
\bibinfo{author}{Rennie, S.~J.}, \bibinfo{author}{Marcheret, E.}, \bibinfo{author}{Mroueh, Y.}, \bibinfo{author}{Ross, J.} \& \bibinfo{author}{Goel, V.}
\newblock \bibinfo{title}{Self-critical sequence training for image captioning}.
\newblock In \emph{\bibinfo{booktitle}{CVPR}}, \bibinfo{pages}{1179--1195} (\bibinfo{year}{2017}).

\bibitem{lu2017knowing}
\bibinfo{author}{Lu, J.}, \bibinfo{author}{Xiong, C.}, \bibinfo{author}{Parikh, D.} \& \bibinfo{author}{Socher, R.}
\newblock \bibinfo{title}{Knowing when to look: Adaptive attention via a visual sentinel for image captioning}.
\newblock In \emph{\bibinfo{booktitle}{CVPR}}, \bibinfo{pages}{3242--3250} (\bibinfo{year}{2017}).

\bibitem{anderson2018bottomup}
\bibinfo{author}{Anderson, P.} \emph{et~al.}
\newblock \bibinfo{title}{Bottom-up and top-down attention for image captioning and visual question answering}.
\newblock In \emph{\bibinfo{booktitle}{CVPR}}, \bibinfo{pages}{6077--6086} (\bibinfo{year}{2018}).

\bibitem{liu-etal-2021-contrastive}
\bibinfo{author}{Liu, F.} \emph{et~al.}
\newblock \bibinfo{title}{Contrastive attention for automatic chest {X}-ray report generation}.
\newblock In \emph{\bibinfo{booktitle}{ACL-IJCNLP}}, \bibinfo{pages}{269--280} (\bibinfo{year}{2021}).

\bibitem{liu-etal-2021-competence}
\bibinfo{author}{Liu, F.}, \bibinfo{author}{Ge, S.} \& \bibinfo{author}{Wu, X.}
\newblock \bibinfo{title}{Competence-based multimodal curriculum learning for medical report generation}.
\newblock In \emph{\bibinfo{booktitle}{ACL-IJCNLP}}, \bibinfo{pages}{3001--3012} (\bibinfo{year}{2021}).

\bibitem{liu2021ppked}
\bibinfo{author}{Liu, F.}, \bibinfo{author}{Wu, X.}, \bibinfo{author}{Ge, S.}, \bibinfo{author}{Fan, W.} \& \bibinfo{author}{Zou, Y.}
\newblock \bibinfo{title}{Exploring and distilling posterior and prior knowledge for radiology report generation}.
\newblock In \emph{\bibinfo{booktitle}{CVPR}}, \bibinfo{pages}{13748--13757} (\bibinfo{year}{2021}).

\bibitem{chen-etal-2021-cross-modal}
\bibinfo{author}{Chen, Z.}, \bibinfo{author}{Shen, Y.}, \bibinfo{author}{Song, Y.} \& \bibinfo{author}{Wan, X.}
\newblock \bibinfo{title}{Cross-modal memory networks for radiology report generation}.
\newblock In \emph{\bibinfo{booktitle}{ACL-IJCNLP}}, \bibinfo{pages}{5904--5914} (\bibinfo{year}{2021}).

\bibitem{qin-song-2022-reinforced}
\bibinfo{author}{Qin, H.} \& \bibinfo{author}{Song, Y.}
\newblock \bibinfo{title}{Reinforced cross-modal alignment for radiology report generation}.
\newblock In \emph{\bibinfo{booktitle}{ACL}}, \bibinfo{pages}{448--458} (\bibinfo{year}{2022}).

\bibitem{yang2023radiology}
\bibinfo{author}{Yang, S.} \emph{et~al.}
\newblock \bibinfo{journal}{\bibinfo{title}{Radiology report generation with a learned knowledge base and multi-modal alignment}}.
\newblock {\emph{\JournalTitle{Med. Image Anal.}}} \textbf{\bibinfo{volume}{86}}, \bibinfo{pages}{102798} (\bibinfo{year}{2023}).

\bibitem{li2023dynamic}
\bibinfo{author}{Li, M.} \emph{et~al.}
\newblock \bibinfo{title}{Dynamic graph enhanced contrastive learning for chest x-ray report generation}.
\newblock In \emph{\bibinfo{booktitle}{CVPR}}, \bibinfo{pages}{3334--3343} (\bibinfo{year}{2023}).

\bibitem{tang2025dual}
\bibinfo{author}{Tang, Q.} \emph{et~al.}
\newblock \bibinfo{journal}{\bibinfo{title}{Dual-modality visual feature flow for medical report generation}}.
\newblock {\emph{\JournalTitle{Med. Image Anal.}}} \textbf{\bibinfo{volume}{101}}, \bibinfo{pages}{103413} (\bibinfo{year}{2025}).

\bibitem{tanida2023interactive}
\bibinfo{author}{Tanida, T.}, \bibinfo{author}{Müller, P.}, \bibinfo{author}{Kaissis, G.} \& \bibinfo{author}{Rueckert, D.}
\newblock \bibinfo{title}{{Interactive and Explainable Region-guided Radiology Report Generation}}.
\newblock In \emph{\bibinfo{booktitle}{CVPR}}, \bibinfo{pages}{7433--7442} (\bibinfo{year}{2023}).

\bibitem{nicolson2023improving}
\bibinfo{author}{Nicolson, A.}, \bibinfo{author}{Dowling, J.} \& \bibinfo{author}{Koopman, B.}
\newblock \bibinfo{journal}{\bibinfo{title}{Improving chest x-ray report generation by leveraging warm starting}}.
\newblock {\emph{\JournalTitle{Artif. Intell. Med.}}} \textbf{\bibinfo{volume}{144}}, \bibinfo{pages}{102633} (\bibinfo{year}{2023}).

\bibitem{gao2025s2d}
\bibinfo{author}{Gao, J.}, \bibinfo{author}{Liu, C.} \& \bibinfo{author}{Li, Y.}
\newblock \bibinfo{journal}{\bibinfo{title}{S2d-align: Shallow-to-deep auxiliary learning for anatomically-grounded radiology report generation}}.
\newblock {\emph{\JournalTitle{arXiv preprint arXiv:2511.11066}}}  (\bibinfo{year}{2025}).

\bibitem{HZR16}
\bibinfo{author}{He, K.}, \bibinfo{author}{Zhang, X.}, \bibinfo{author}{Ren, S.} \& \bibinfo{author}{Sun, J.}
\newblock \bibinfo{title}{Deep residual learning for image recognition}.
\newblock In \emph{\bibinfo{booktitle}{CVPR}}, \bibinfo{pages}{770--778} (\bibinfo{year}{2016}).

\bibitem{DCL19}
\bibinfo{author}{Devlin, J.}, \bibinfo{author}{Chang, M.-W.}, \bibinfo{author}{Lee, K.} \& \bibinfo{author}{Toutanova, K.}
\newblock \bibinfo{title}{Bert: Pre-training of deep bidirectional transformers for language understanding}.
\newblock In \emph{\bibinfo{booktitle}{NAACL-HLT}}, \bibinfo{pages}{4171--4186} (\bibinfo{year}{2019}).

\bibitem{radford2021clip}
\bibinfo{author}{Radford, A.} \emph{et~al.}
\newblock \bibinfo{title}{Learning transferable visual models from natural language supervision}.
\newblock In \emph{\bibinfo{booktitle}{International Conference on Machine Learning}}, vol. \bibinfo{volume}{139}, \bibinfo{pages}{8748--8763} (\bibinfo{year}{2021}).

\bibitem{gupta2021memory}
\bibinfo{author}{Gupta, A.}, \bibinfo{author}{Dar, G.}, \bibinfo{author}{Goodman, S.}, \bibinfo{author}{Ciprut, D.} \& \bibinfo{author}{Berant, J.}
\newblock \bibinfo{title}{Memory-efficient transformers via top-k attention}.
\newblock In \emph{\bibinfo{booktitle}{EMNLP}}, \bibinfo{pages}{39--52} (\bibinfo{year}{2021}).

\bibitem{yuan2025native}
\bibinfo{author}{Yuan, J.} \emph{et~al.}
\newblock \bibinfo{title}{Native sparse attention: Hardware-aligned and natively trainable sparse attention}.
\newblock In \emph{\bibinfo{booktitle}{ACL}}, \bibinfo{pages}{23078--23097} (\bibinfo{year}{2025}).

\bibitem{lin2017focal}
\bibinfo{author}{Lin, T.-Y.}, \bibinfo{author}{Goyal, P.}, \bibinfo{author}{Girshick, R.}, \bibinfo{author}{He, K.} \& \bibinfo{author}{Doll{\'a}r, P.}
\newblock \bibinfo{title}{Focal loss for dense object detection}.
\newblock In \emph{\bibinfo{booktitle}{IEEE International Conference on Computer Vision}}, \bibinfo{pages}{2980--2988} (\bibinfo{year}{2017}).

\bibitem{kendall2018multi}
\bibinfo{author}{Kendall, A.}, \bibinfo{author}{Gal, Y.} \& \bibinfo{author}{Cipolla, R.}
\newblock \bibinfo{title}{Multi-task learning using uncertainty to weigh losses for scene geometry and semantics}.
\newblock In \emph{\bibinfo{booktitle}{IEEE Conference on Computer Vision and Pattern Recognition}}, \bibinfo{pages}{7482--7491} (\bibinfo{year}{2018}).

\bibitem{PRW02}
\bibinfo{author}{Papineni, K.}, \bibinfo{author}{Roukos, S.}, \bibinfo{author}{Ward, T.} \& \bibinfo{author}{Zhu, W.-J.}
\newblock \bibinfo{title}{Bleu: a method for automatic evaluation of machine translation}.
\newblock In \emph{\bibinfo{booktitle}{ACL}}, \bibinfo{pages}{311--318} (\bibinfo{year}{2002}).

\bibitem{DeL11}
\bibinfo{author}{Denkowski, M.} \& \bibinfo{author}{Lavie, A.}
\newblock \bibinfo{title}{Meteor 1.3: Automatic metric for reliable optimization and evaluation of machine translation systems}.
\newblock In \emph{\bibinfo{booktitle}{ACL}}, \bibinfo{pages}{85--91} (\bibinfo{year}{2011}).

\bibitem{Lin04}
\bibinfo{author}{Lin, C.-Y.}
\newblock \bibinfo{title}{Rouge: A package for automatic evaluation of summaries}.
\newblock In \emph{\bibinfo{booktitle}{Text summarization branches out}}, \bibinfo{pages}{74--81} (\bibinfo{year}{2004}).

\bibitem{vedantam2015cider}
\bibinfo{author}{Vedantam, R.}, \bibinfo{author}{Lawrence~Zitnick, C.} \& \bibinfo{author}{Parikh, D.}
\newblock \bibinfo{title}{Cider: Consensus-based image description evaluation}.
\newblock In \emph{\bibinfo{booktitle}{CVPR}}, \bibinfo{pages}{4566--4575} (\bibinfo{year}{2015}).

\bibitem{SJR20}
\bibinfo{author}{Smit, A.} \emph{et~al.}
\newblock \bibinfo{title}{Combining automatic labelers and expert annotations for accurate radiology report labeling using bert}.
\newblock In \emph{\bibinfo{booktitle}{EMNLP}}, \bibinfo{pages}{1500--1519} (\bibinfo{year}{2020}).

\bibitem{jain2021radgraph}
\bibinfo{author}{Jain, S.} \emph{et~al.}
\newblock \bibinfo{title}{{RadGraph}: Extracting clinical entities and relations from radiology reports}.
\newblock In \emph{\bibinfo{booktitle}{Neural Information Processing Systems Track on Datasets and Benchmarks}} (\bibinfo{year}{2021}).

\end{thebibliography}

\end{document}